\documentclass[letterpaper, 10 pt, conference]{ieeeconf}  

\IEEEoverridecommandlockouts                              

\usepackage{graphics} 
\usepackage{booktabs}
\usepackage{subfigure}
\usepackage{amsmath,amssymb,amsfonts}
\usepackage{cite}
\usepackage{tabulary}
\usepackage{tabularray}
\usepackage{tikz}
\usepackage{color}

\title{\LARGE \bf
OpenOcc: Open Vocabulary 3D Scene Reconstruction via Occupancy Representation
}

\author{Haochen Jiang$^{1*}$, Yueming Xu$^{1*}$, Yihan Zeng$^{2}$, Hang Xu$^{2}$, Wei Zhang$^{2}$, Jianfeng Feng$^{1}$, Li Zhang$^{1\dag}$
\thanks{This work is supported by National Key R\&D Program of China (No.2019YFA0709502, No.2018YFC1312904), National Natural Science Foundation of China (Grant No. 62106050 and 62376060), Natural Science Foundation of Shanghai (Grant No. 22ZR1407500), ZJ Lab, and Shanghai Center for Brain Science and Brain-Inspired Technology.}
\thanks{*Haochen Jiang and Yueming Xu contributed equally to this work.}
\thanks{$^{1}$Haochen Jiang, Yueming Xu, Jianfeng Feng, and Li Zhang are with Fudan University. (\dag Li Zhang is the corresponding author: lizhangfd@fudan.edu.cn)}%
\thanks{$^{2}$Yihan Zeng, Hang Xu and Wei Zhang are with Huawei Noah’s Ark Lab.}
\thanks{$^3$https://github.com/fudan-zvg/OpenOcc}
}

\makeatletter
\renewcommand{\paragraph}{%
  \@startsection{paragraph}{4}%
  {\z@}{1.1ex \@plus 1ex \@minus .2ex}{-1mm}%
  {\normalfont\normalsize\bfseries}%
}
\makeatother

\begin{document}
\maketitle
\thispagestyle{empty}
\pagestyle{empty}

\begin{abstract}
3D reconstruction has been widely used in autonomous navigation fields of mobile robotics. However, the former research can only provide the basic geometry structure without the capability of open-world scene understanding, limiting advanced tasks like human interaction and visual navigation. Moreover, traditional 3D scene understanding approaches rely on expensive labeled 3D datasets to train a model for a single task with supervision.
Thus, geometric reconstruction with zero-shot scene understanding i.e. Open vocabulary 3D Understanding and Reconstruction, is crucial for the future development of mobile robots. 
In this paper, we propose OpenOcc, a novel framework unifying the 3D scene reconstruction and open vocabulary understanding with neural radiance fields. We model the geometric structure of the scene with occupancy representation and distill the pre-trained open vocabulary model into a 3D language field via volume rendering for zero-shot inference. 
Furthermore, a novel semantic-aware confidence propagation (SCP) method has been proposed to relieve the issue of language field representation degeneracy caused by inconsistent measurements in distilled features.
Experimental results show that our approach achieves competitive performance in 3D scene understanding tasks, especially for small and long-tail objects.
\end{abstract}    
\section{Introduction}
\label{sec:intro}
\begin{figure}[!t]
\centering
\includegraphics[width=\columnwidth]{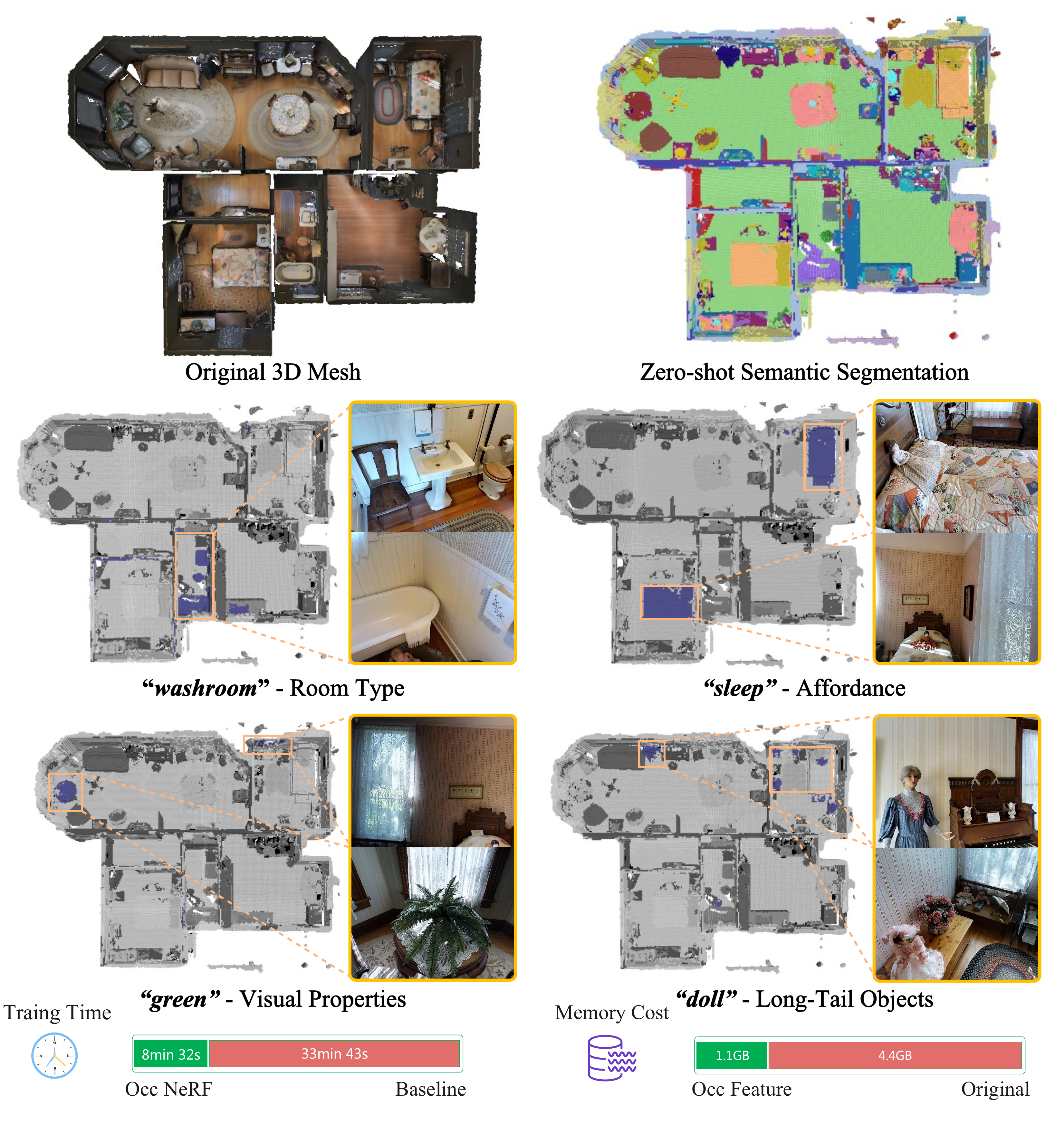} 
\caption{\textbf{Open-vocabulary 3D scene understanding and reconstruction.} We propose OpenOcc, a zero-shot method for 3D scene structure perception. The examples above show zero-shot 3D scene semantic segmentation results with the occupancy feature grid. The blue color denotes the matching results of a user-specified query string to demonstrate the flexibility of the language-based features grid.}
\label{fg:intro query show}
\vspace{-3mm}
\end{figure}

Reconstructing 3D scenes with high-level semantics from digital images or videos is a crucial research area in 3D computer vision and robotics, which has been widely used in various fields such as visual navigation~\cite{galindo2008robot,huang2023visual}, and robotic grasping~\cite{antanas2019semantic,shen2023F3RM}. However, the intrinsic connections and underlying homologies between scene reconstruction and understanding have been neglected. These two tasks are often treated as separate studies, where the bottleneck lies. 

Several commonly used 3D representations include 3D point clouds~\cite{qi2017pointnet,qi2017pointnet++}, voxel grids~\cite{hu2021vmnet,meng2019vv}, and polygonal meshes~\cite{hanocka2019meshcnn}. Directly learning semantics on these 3D representations requires a substantial amount of well-annotated 3D ground truths for training. It is more laborious and expensive to obtain the 3D ground truth label. Moreover, the majority of existing methods~\cite{sematnicnerfICCV2021, wei2023surroundocc,zhang2023occformer,huang2023tri} encounter challenges in achieving model transferability owing to domain gaps. Recently, neural radiance fields (NeRFs) \cite{mildenhall2021nerf} have demonstrated the power of implicit representation in intricate real-world 3D scenes, which can capture intricate geometric details using solely RGB images. This novel scene representation is more compact, continuous, efficient, and able to be optimized with differentiable rendering. Many researchers utilize this 3D representation to conduct the Neural Semantic fields~\cite{sematnicnerfICCV2021,kundu2022panoptic}. However, these existing works do not incorporate geometric (depth) and multi-view constraints during the training process, leading to the emergence of shape-radiance ambiguity like LeRF~\cite{kerr2023lerf}. Moreover, these methods can not effectively reconstruct geometric structures in large-scale scenes, like Matport3D~\cite{Matterport3D_2017}, and lack the open vocabulary capabilities to understand environments. Most reconstruction works utilize this 3D representation~\cite {mescheder2019occupancy,peng2020convolutional} to achieve high-fidelity and multi-view consistent reconstructions with reduced memory usage. However, the reconstructed geometric structure lacks the incorporation of high-level semantics. 
Hence, we ask a natural question: 

\textit{Can we customize a better way to effectively integrate the benefits of scene reconstruction and open vocabulary 3D scene understanding for generating a consistent semantic map?}

However, it is not feasible to directly fuse these two types of work for the robotic structure perception, owing to two facts: 1) Current reconstruction methods~\cite{mescheder2019occupancy, oechsle2021unisurf, yang2022s} only perform better on 3D object surface reconstruction, which can not be extended to complex or large-scale indoor scenes. 2) Owing to the potential noise of language embedding, the open vocabulary segmentation predictions are ambiguous. Directly distilling these inconsistent features may lead to the degeneracy of the representation capabilities for language fields, which causes floaters in volume rendering. 

To tackle this question, we propose \textit{OpenOcc}, a novel framework unifying the scene reconstruction and open vocabulary understanding utilizing NeRF. To better learn the implicit representation, we construct the language, geometric, and appearance fields with separate multi-resolution feature gird. Instead of predicting a sign distance function (SDF), we utilize the occupancy representation surface as the decision boundary of a binary occupancy classifier. This leads to a significant reduction in memory requirements and improves the inference efficiency compared to the dense point-based and mesh-based representation. Moreover, the small objects and thin structures can be better reconstructed. To achieve open vocabulary scene understanding, we distill the 2D language embeddings into a 3D language field inspired by LeRF~\cite{kerr2023lerf}. To address the issue of shape-radiance ambiguity, we have developed geometric constraints that incorporate depth information. Owing to the potential noise in the language embedding, the predictions from open vocabulary segmentation exhibit inconsistencies and ambiguities. Thus, we propose an SCP method to dynamically reallocate the weights of the corresponding language features in the language field updating, which can significantly improve the accuracy of 2D segmentation results.

In summary, our \textbf{contributions} are summarized as follows:
\begin{itemize}
    \item We design a novel open vocabulary reconstruction system, unifying the 3D scene reconstruction and understanding with neural radiance fields, which can simultaneously achieve geometric reconstruction and zero-shot perception from a set of posed RGB-D images.
    \item We introduce a novel approach of semantic-aware confidence propagation (SCP) within the process of updating semantic radiance field features. This method addresses the challenge of inconsistent open vocabulary segmentation across diverse viewpoints.
    \item We demonstrate that the proposed methods can be used for 3D semantic segmentation with better performance than state-of-the-art zero-shot methods, particularly for small and long-tail objects.
\end{itemize}

\begin{figure*}[t]
\centering
\includegraphics[width=0.93\textwidth]{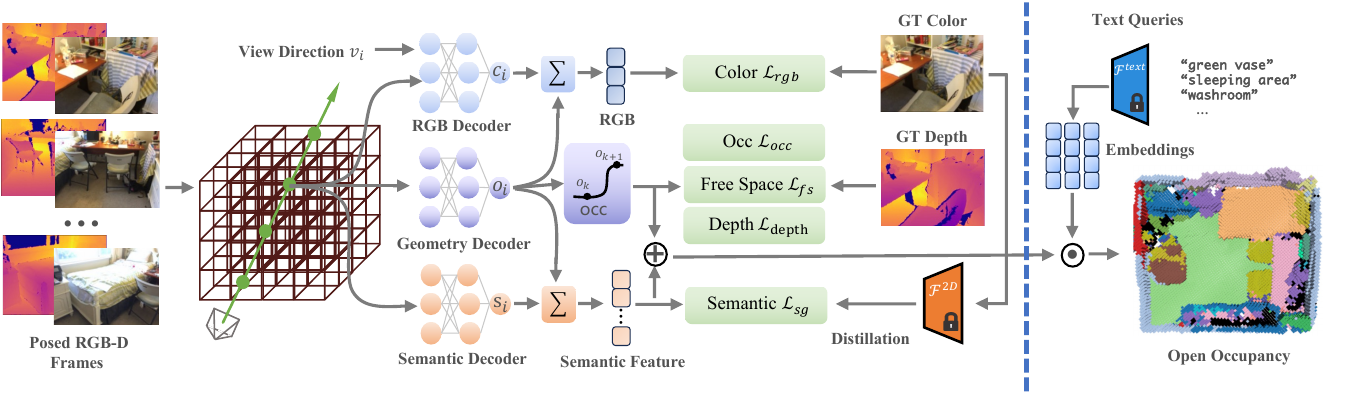} 
\vspace{-2mm}
\caption{\textbf{The overall framework of the proposed method.} \textit{Left:} Given a series of posed RGB-D frames, we construct the RGB, geometry, and semantic decoders via separate multi-resolution feature gird with the geometric loss $\mathcal{L}_{occ}, \mathcal{L}_{fs}, \mathcal{L}_{depth}$ and color loss $\mathcal{L}_{color}$. To learn the language knowledge, we distill the dense feature $\mathcal{F}_{2d}$ via volume rendering with a distillation loss $\mathcal{L}_{sg}$. \textit{Right:} During inference, we can compute the similarity score between user's text embeddings and generate an occupancy feature map to perform the open-vocabulary 3D understanding task.}
\label{fg:pipeline}
\vspace{-3mm}
\end{figure*}

\section{Related Work}
\label{sec:related}

\paragraph{Nerual Implicit 3D Scene Reconstruction}
Neural implicit scene representations, also known as neural fields~\cite{mildenhall2021nerf}, have gained considerable attention in the field of 3D geometry reconstruction owing to their high level of expressivity and minimal memory requirements. NeuS~\cite{wang2021neus} introduces a volume rendering method to train an unbiased neural Signed Distance Function (SDF) representation. VolSDF~\cite{yariv2021volume} proposes a parameterization method to transfer the volume density to SDF. UniSurf~\cite{oechsle2021unisurf} unifies neural volume and surface rendering, enabling both within the same model. These works only perform better on 3D object reconstruction, which can not expand to complex indoor scenes. Thus, there occurs some works~\cite{MonoSDF,azinovic2022neural}, exploiting geometric priors to improve
multi-view 3D scene reconstruction quality, efficiency, and scalability. Many works follow this framework and improve it in many aspects, such as model acceleration~\cite{wang2022go, Fast_Monocular}, and reconstruction precision~\cite{ResNerf, guo2022manhattan, Occ-SDF}. Considering the critical factors of efficiency, reconstruction quality, and integrity of 3D scene depiction for robotic navigation, we exploit a more practical occupancy representation instead of SDF to reconstruct the indoor scenes inspired by~\cite{mescheder2019occupancy, peng2020convolutional}. 

\paragraph{Open-Vocabulary 2D Scene Understanding} Visual Language Models (VLMs)~\cite{radford2021learning,jia2021scaling} have recently achieved the capability to establish robust mappings between images and textual content, leading to a remarkable level of performance in zero-shot 2D scene understanding tasks. However, VLMs only provide image-level embeddings, which can not apply to the dense prediction task requiring pixel-level information such as segmentation. Thus, many works~\cite{liu2022open,li2022languagedriven,ghiasi2022scaling,xu2023learning,liang2023open} attempt to align the per-pixel image feature with the large language embeddings. In this way, users can give an arbitrary text label and query the interest objects on-the-fly. Compared with the close-set 2D scene understanding methods~\cite{wu2022p2t,zhou2023bcinet}, we exploit a pre-trained open-vocabulary 2D models~\cite{liu2022open,li2022languagedriven} and distill the 2D feature into 3D to finish a series of 3D understanding tasks, without any ground truth label in the whole training process. Leveraging the robust comprehension power of an open vocabulary, our approach demonstrates proficiency in recognizing long-tail objects and those objects that lack label annotations within the dataset. This effectively facilitates the open-ended perception potential crucial for autonomous robotic systems.

\paragraph{Grounding Language into 3D Representations}
Neural implicit feature fields have emerged as a method to translate 2D features into a grounded 3D representation. CLIP-Fields~\cite{shafiullah2022clip} fuse CLIP~\cite{radford2021learning} embeddings of crops into point clouds using a contrastively supervised field. OpenScene~\cite{peng2023openscene} and VL-Maps~\cite{huang2023visual} build a 3D volume of language features aligned with the pre-trained language segmentation encoders~\cite{li2022languagedriven, ghiasi2022scaling}. These methods are significantly constrained as they do not encode the intrinsic scene geometry, relying instead on an external point cloud for query operations. LERF~\cite{kerr2023lerf} and VL-Fields~\cite{tsagkas2023vlfields} advocate the construction of a language field through volume rendering and the incorporation of the 2D language feature~\cite{radford2021learning, li2022languagedriven} into the 3D implicit field, which is beneficial to obtain the dense and continuous scene perception. However, owing to the potentially noisy nature of language embeddings, the predictions from open vocabulary segmentation exhibit inconsistencies and ambiguity. In this paper, we propose a semantic-aware confidence propagation method to dynamically reallocate the weights of the corresponding language features in the language field updating. The mistake observation of language embeddings can be efficiently filtered.

\section{Methodology}
\label{sec:method}

The overall architecture of the proposed methods is illustrated in Fig.~\ref{fg:pipeline}. The entire framework can be divided into three parts. Firstly, an explicit occupancy neural radiance field is trained to reconstruct the geometric structure of the environments. Secondly, we introduce a semantic language field (SLF) utilizing the open-vocabulary semantic segmentation methods such as OpenSeg~\cite{ghiasi2022scaling} to distill the 2D features into 3D space, which improves the robotics ability of 3D scene understanding. Finally, during inference, we can achieve an open vocabulary segmentation task by computing the similarity scores between an arbitrary text query and a generated semantic feature map with occupancy representation.

\subsection{Occupancy Field Generation}
\paragraph{Multi-resolution Feature Grid} 
To better reconstruct details of the 3D scene geometry structure, we utilize the multi-resolution feature grid~\cite{muller2022instant} via occupancy representation instead of a single scale inspired by Go-Surf~\cite{wang2022go}. We encode the scene geometry into a multi-level feature grid $\mathbb{V}_{\theta} = \{V^l\}$, where $l \in \{0, 1, 2, ..., L\}$ encode the geometry details from coarse to fine. The coarse-level features can fastly initialize scene geometry presentation to fill the hole. Furthermore, the geometry residual of the high-frequency details will be refined by the fine-level features grid. 

For a random sample point $\mathbf{p} \in \mathbb{R}^3$ in a scene, we query the representation feature by tri-linear interpolation operation at each level and concat the multi-level features as the input of the decoder to obtain the occupancy probability $o(p)$ via the MLP $f_{\alpha}(\cdot)$:
\begin{equation}
\mathcal{V}_\theta(\mathbf{p}) =\left[V^0(\mathbf{p}), V^1(\mathbf{p}), V^2(\mathbf{p}),...,V^L(\mathbf{p})\right],
\end{equation}
\begin{equation}
    o(\mathbf{p}) =f_\alpha\left(\mathcal{V}_\theta(\mathbf{p})\right).
\end{equation}
To reduce the ambiguity between geometry and appearance in the learning process, we use a separate feature grid $\mathcal{C}_\phi, \mathcal{S}_\omega$ to module the color and semantic fields, respectively. Similarly, we use different decoders $g_{\beta}(\cdot), h_{\gamma}(\cdot)$ to infer the color and semantic features.
\begin{equation}
\label{eq:color decoder}
c(\mathbf{p}, \mathbf{d})=g_\beta\left(\mathcal{C}_\phi(\mathbf{p}), \mathbf{d}\right),
\end{equation}
\begin{equation}
\label{eq:semantic decoder}
s(\mathbf{p})=h_\gamma\left(\mathcal{S}_\omega(\mathbf{p})\right),
\end{equation}
where $\mathbf{d}$ is the view direction of the sample ray. Here $\theta, \phi, \omega$ and $\alpha, \beta, \gamma$ represent the optimization parameters about grid features and MLP weights, respectively.

\paragraph{Volume Rendering with Occupancy Field} 
To obtain the explicit feature grid indirectly, we adopt an occupancy field instead of a density field to represent the scene's geometry structure inspired by \cite{mescheder2019occupancy} and \cite{peng2020convolutional}. Instead of predicting a discrete voxel representation, we predict the continuous occupancy function to generate an occupancy field which is similar to SDF representation. Given a ray $\mathbf{r}$ from the virtual camera's center connected with a random pixel $[u,v]$, we sample a set of $N$ 3D points $\{\mathbf{x}_i \in \mathbb{R}^3\}_N$ along the viewing direction $\mathbf{d} \in \mathbb{R}^3$. Following the Eq.~\ref{eq:color decoder}, we can obtain the predicted RGB color value $c_i(\mathbf{x}, \mathbf{d}) \in \mathbb{R}^3$ at each position $\mathbf{x}$. For every location $\mathbf{x}$, we predict an occupancy probability between 0 and 1 via a neural network as Eq.~\ref{eq:occ map}. 
\begin{equation}
\label{eq:occ map}
    o(\mathbf{p}) : \mathbb{R}^3 \longrightarrow \{0, 1\}
\end{equation}
where $o(\mathbf{p})=0$ and $o(\mathbf{p})=1$ denotes the points located in free space and occupied space, respectively. Note that we are interested in the decision boundary, which implicitly represents the scene's surface with a continuous occupancy field. According to the physical implications of accumulated transmittance $T_i =\exp \left(-\sum_{j<i} \sigma\left(\mathbf{x}_j\right) \delta_j\right)$, the rendering equation of color and depth can be rewritten as : 
\begin{equation}
\label{eq:occ color render}
\resizebox{0.31\textwidth}{!}{$
\begin{aligned}
    \hat{C}(\mathbf{r}) &=\sum_{i=1}^N o(\mathbf{x}_i) \prod_{j<i}(1-o(\mathbf{x}_j)) c\left(\mathbf{x}_i, \mathbf{d}\right), \\
    \hat{D}(\mathbf{r}) &=\sum_{i=1}^N o(\mathbf{x}_i) \prod_{j<i}(1-o(\mathbf{x}_j)) z_i,
\end{aligned}
$}
\end{equation}
where the terms $\prod_{j<i}(1-o(\mathbf{x}_j))$ denotes the visible probability which is 1 if there is no occupied sample $\mathbf{x}_j$ with $j<i$ before sample $\mathbf{x}_i$. Moreover, $o(\mathbf{x}_i)\prod_{j<i}(1-o(\mathbf{x}_j))$ represents the ray first hit the surface at $\mathbf{x}_i$ which the occupied value is 1, and other samples is 0. Thus, $\hat{C}(\mathbf{r})$ and $\hat{D}(\mathbf{r})$ takes the color $c(\mathbf{x}_i, \mathbf{d})$, depth $z(\mathbf{x}_i)$ of the first occupied sample $\mathbf{x}_i$ along the ray, respectively. 
\subsection{Multi-view Consistent Language Fields}
\paragraph{Image Feature Extraction} 
Given an RGB image $\mathbf{I}_i$ with a resolution of H $\times$ W, we can obtain the per-pixel embedding from a pre-trained visual language segmentation model $\mathcal{F}(\cdot)$ as :
\begin{equation}
     \mathcal{F}(k_i): \mathbb{R}^3 \longrightarrow \mathbb{R}^D,  k_i \in I[u, v]  
\end{equation}
where $\mathcal{F}(\mathbf{I}_i) \in \mathbb{R}^{H \times W \times D}$, $D$ is the language feature dimension, and $i$ is the index sample from a image sequences. 

\begin{figure}[t]
\centering
\includegraphics[width=0.9\columnwidth]{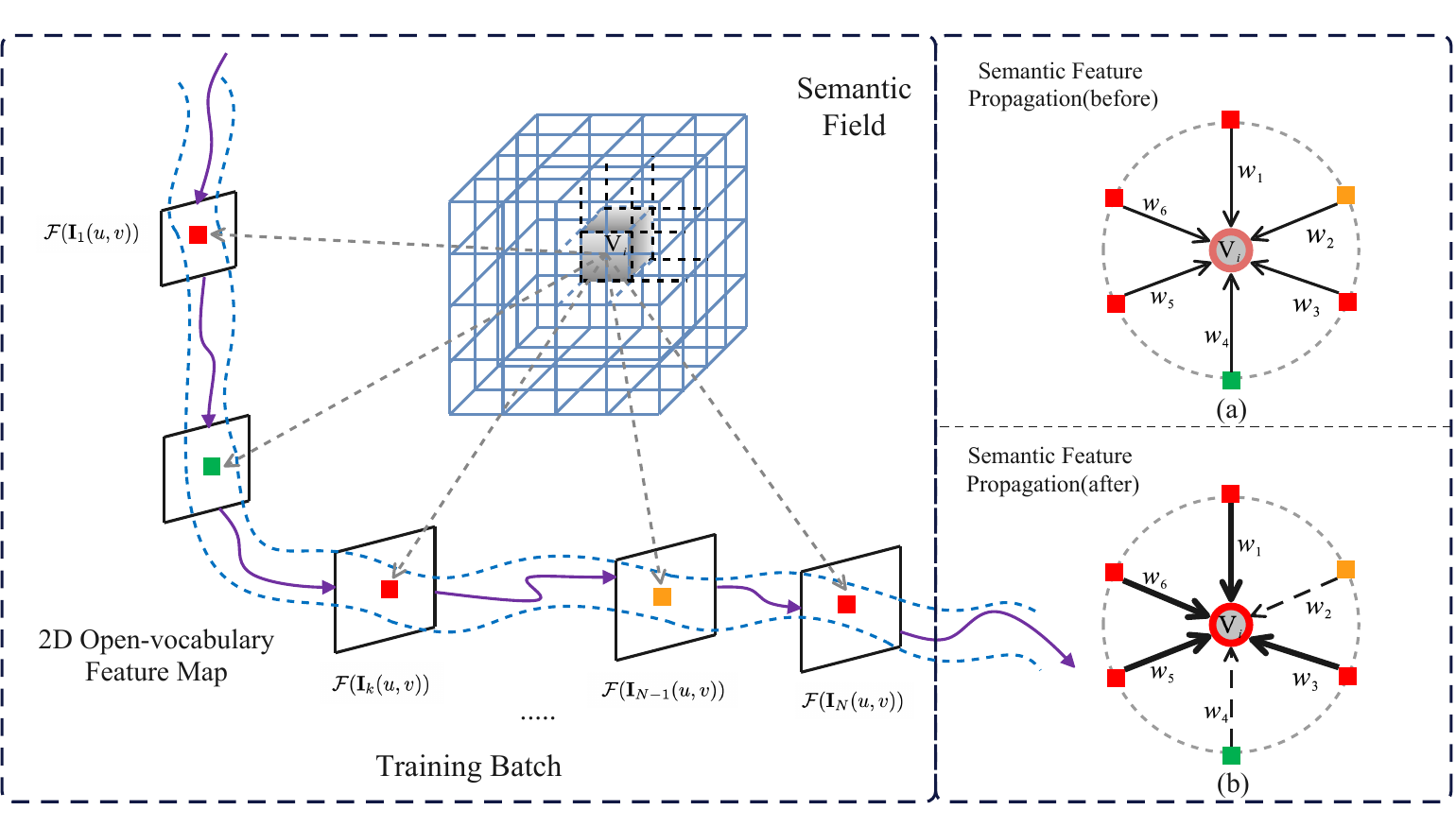} 
\caption{\textbf{Different semantic feature update strategies.} Owing to the potentially noisy language embedding, the open vocabulary segmentation results in different views are inconsistent in the same training batch that can show the left image. (a) Semantic field updates use the same weight. (b) The proposed Semantic-aware Confidence Propagation (SCP). Dashed lines mean smaller weight. The boundary color of the center point means the fusing feature is dominated by which consistent semantic class.}
\label{fg:semantic confidence propagation}
\vspace{-3mm}
\end{figure}

\paragraph{Features Distillation} 
To obtain the queryable scene representation, we distill the 2D visual-language feature into 3D explicit semantic fields utilizing volume rendering inspired by \cite{sematnicnerfICCV2021, peng2023openscene, kerr2023lerf}. Given a position $\mathbf{x}_i$ in space, we can query the volume feature and decode the semantic feature as Eq.~\ref{eq:semantic decoder}. Similarly, the expected semantic feature $\hat{S}(\mathbf{r})$ along each pixel/ray can be defined as :
\begin{equation}
\hat{S}(\mathbf{r}) =\sum_{i=1}^N o(\mathbf{x}_i) \prod_{j<i}(1-o(\mathbf{x}_j))s(\mathbf{x}_i),
\end{equation}
where the dimension of $\hat{S}(\mathbf{r})$ is $D$, which is equal and corresponding to $\mathcal{F}(k_i)$. To keep consistent with the CLIP feature, we normalize each raw output to the unit sphere as
$S(\mathbf{r}) = \hat{S}(\mathbf{r}) / {\Vert \hat{S}(\mathbf{r}) \Vert}$.
Since the open-vocabulary segmentation embedding from \cite{ghiasi2022scaling, li2022languagedriven} is co-embedded with CLIP feature, the distilled 3D semantic language field is also located in the same space as CLIP naturally. 

\paragraph{Semantic-aware Confidence Propagation (SCP)}
\label{subsec:SCP}
To obtain a more consistent and accurate semantic representation within the language field, we propose a probabilistic fusion approach, denoted as SCP, to integrate the multi-view semantic features as shown in Fig.~\ref{fg:semantic confidence propagation}, which can dynamically adjust the weights of the features during the language field updating. 
During a discrete update interval at time $t$, each feature grid is associated with $N$ measurements in a batch, represented as $z_{t}(N)$. Denote the k-th label probability distribution of a feature grid at the current optimization step as $s_t(k)$, we consider the history sampling measurements till now as $z_{1:t}$ and then update the 3D semantic language fields based on the Bayes' rule:
\begin{equation}
\label{eq:posterior}
P\left(s_t \mid z_{1: t}\right) = \frac{P\left(s_t \mid z_t\right) P\left(z_t\right)}{P\left(s_t\right)} \frac{P\left(s_{t-1} \mid z_{1: t-1}\right)}{P\left(z_t \mid z_{1: t-1}\right)}.
\end{equation}
The opposite event $\tilde{s}_t$ of posterior distribution can be obtained by:
\begin{equation}
\label{eq:posterior_hat}
\resizebox{0.49\textwidth}{!}{$
P\left(\tilde{s}_t \mid z_{1:t}\right)=\frac{\left(1-P\left(s_t \mid z_t\right)\right) P\left(z_t\right)\left(1-P\left(s_{t-1} \mid z_{1: t-1}\right)\right)}{(1-P(s_t)) P\left(z_t \mid z_{1: t-1}\right)}.
$}
\end{equation}
Dividing the above two equations Eq.~\ref{eq:posterior} and Eq.~\ref{eq:posterior_hat}.
We utilize the log-odds algorithm from Thrun~\cite{thrun2002probabilistic} to quantify the update weights for semantic labels:
\begin{equation}
l_t(s)=l\left(s_t \mid z_t\right)+l\left(s_{t-1} \mid z_{1: t-1}\right)-l_{0},
\end{equation}
where $l_t(s)=\log\frac{P\left(s_t \mid z_{1: t}\right)}{1-P\left(s_t \mid z_{1: t}\right)}$ denotes the probability of the semantic class at time $t$. 
All semantic classes are initially assigned equal probabilities in each feature grid, which indicates $l_0=0$. During the practical computation process, at each time step $t$, the state $s_t$ typically has $k$ semantic classes in different datasets. 
The observation likelihood for the k-th class, $P\left(s_t(k) \mid z_t\right)$, is calculated as the ratio $N_k/N$. Furthermore, the normalized confidence weight $w_k$ for each semantic label is derived by corresponding $l_t(s_k)$ as
\begin{equation}
w\left(s_k\right)=\frac{l_t(s_k)}{\sum_{j=1}^K l_t(s_j)}
\end{equation}

\begin{figure*}[!t]
\centering
\includegraphics[width=0.90\linewidth]{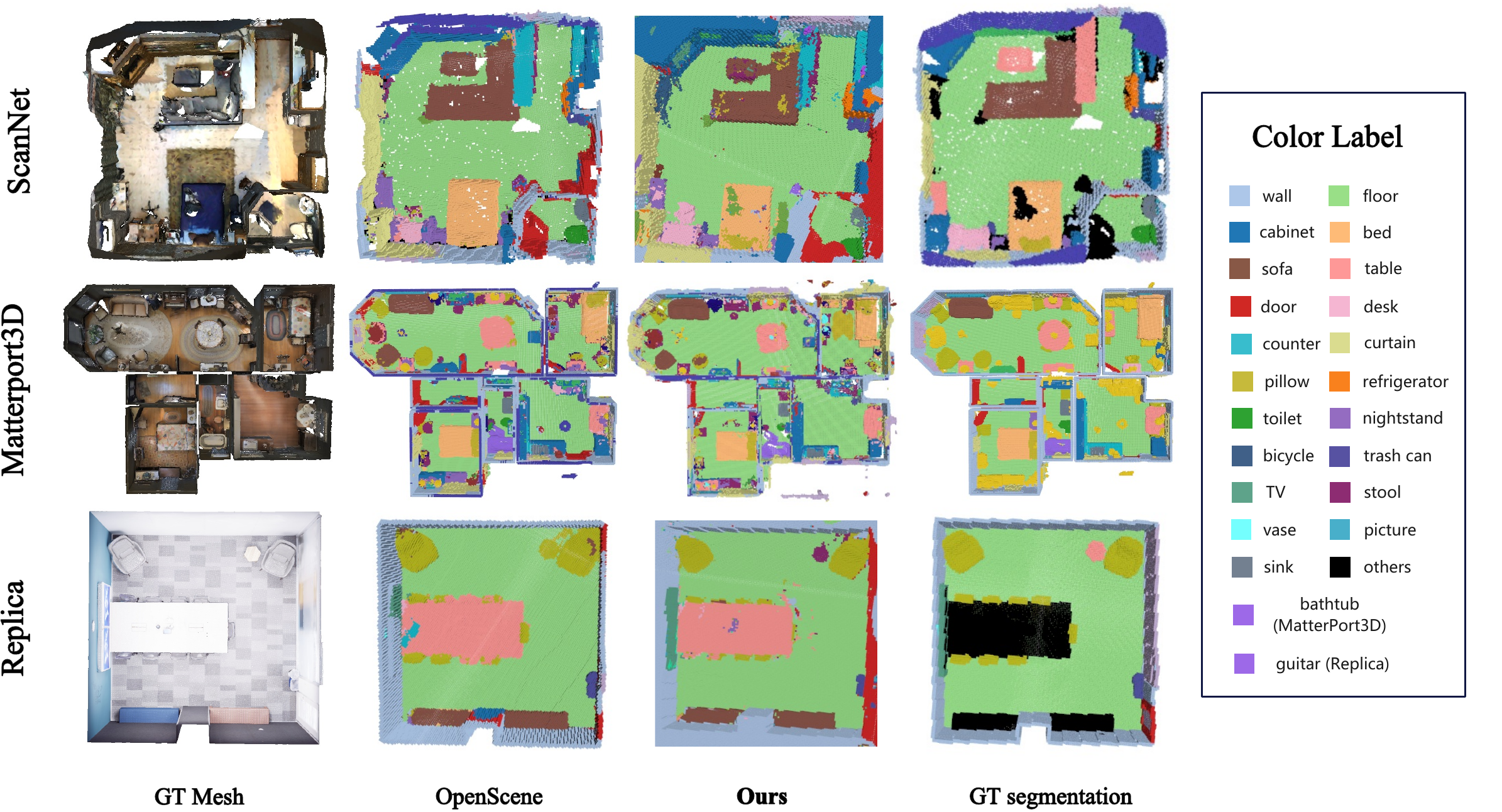}
\vspace{-2mm}
\caption{
\textbf{Qualitative comparisons.} Images of 3D semantic segmentation results on three public indoor benchmarks.
}
\label{fig:occ_seg}
\vspace{-3mm}
\end{figure*}

\subsection{Training Loss}
\paragraph{Color and Depth Rendering Loss} 
To obtain the essential geometry presentation of neural implicit field, we render depth and color in independent view as Eq.~\ref{eq:occ color render} comparing with the proposed ground truth (GT) map: 
\begin{equation}
\begin{aligned}
    \mathcal{L}_{rgb}(\mathbf{r}) &= \frac{1}{M}\sum_{i=1}^M {\| \hat{C}(\mathbf{r}_i) - C(\mathbf{r}_i) \|}^2_2. \\
    \mathcal{L}_{depth}(\mathbf{r}) &= \frac{1}{N_d}\sum_{\mathbf{r} \in N_d} {\| \hat{D}(\mathbf{r}) - D(\mathbf{r}) \|}^2_2.
\end{aligned}
\end{equation}
Where $C(\mathbf{r})$ and $D(\mathbf{r})$ denote the ground truth color and depth corresponding with the given pose, respectively. $M$ represents the number of sampled pixels in the current image. Note that only rays with valid depth value $N_d$ are considered in $\mathcal{L}_{depth}$.

\paragraph{Approximate Occupancy Supervision} 
Directly optimizing the occupancy field from RGB-D observation is not sufficient to reconstruct the high-frequency details and may cause the shape-radiance ambiguity issue. Inspired by SDF constraints in \cite{MonoSDF,wang2022go}, we introduce occupancy supervision with the binary cross-entropy (BCE) loss between the predicted $\hat{o}_{\mathbf{x}_i}$ and the true occupancy values $o_{\mathbf{x}_i}$ as:
\begin{equation}
\label{eq:bce loss}
    \mathcal{L}(\hat{o}_{\mathbf{x}_i}, o_{\mathbf{x}_i}) = -[o_{\mathbf{x}_i}\cdot\log(\hat{o}_{\mathbf{x}_i}) + (1-o_{\mathbf{x}_i})\cdot\log(1-\hat{o}_{\mathbf{x}_i})]
\end{equation}
For the occupancy representation, the surface will exist the thickness owing to the grid resolution. We suppose that the point of occupancy probability, which is located near the surface as $|D(\mathbf{x}_i)-d| \leq t$, is 1. The bound will be controlled by a truncation threshold $t$ (we set $t = 5$ cm). 
\begin{equation}
\label{eq:occ loss}
\resizebox{0.47\textwidth}{!}{$
    \mathcal{L}_{occ}(\hat{o}_{\mathbf{x}_i}, o_{\mathbf{x}_i}) = \frac{1}{M}\sum_{i=1}^M \frac{1}{|\mathcal{R}_{tr}|} \sum_{i \in \mathcal{R}_{tr}} \mathcal{L}(\hat{o}_{\mathbf{x}_i}, o_{\mathbf{x}_i})\bigg|_{o_{\mathbf{x}_i}=1}
$}
\end{equation}
For the depth value out of the truncation, we apply the $\mathcal{L}_{fs}$ loss to encourage the occupancy probability of free space near to 0.
\begin{equation}
\label{eq:fs loss}
\resizebox{0.47\textwidth}{!}{$
    \mathcal{L}_{fs}(\hat{o}_{\mathbf{x}_i}, o_{\mathbf{x}_i}) = \frac{1}{M}\sum_{i=1}^M \frac{1}{|\mathcal{R}_{fs}|} \sum_{i \in \mathcal{R}_{fs}} \mathcal{L}(\hat{o}_{\mathbf{x}_i}, o_{\mathbf{x}_i})\bigg|_{o_{\mathbf{x}_i}=0}
$}
\end{equation}
\paragraph{2D-3D Semantic Feature Aligned Loss} 
To enforce the semantic field aligned with the 2D open vocabulary feature, we use a cosine similarity loss as: 
\begin{equation}
\resizebox{0.43\textwidth}{!}{$
    \mathcal{L}_{sg}(\mathbf{r}) = \frac{1}{M}\sum_{i=1}^M \rho(w_i\cdot{\| 1 -\cos(\mathcal{F}(\mathbf{r}_i), \hat{S}(\mathbf{r}_i)) \|}^2_2)
$}
\end{equation}
where $w_i$ denotes the normalized confidence weight of the corresponding semantic feature, which has been mentioned in Sec. III-B. In addition, $\rho(\cdot)$ is a robust kernel function to restrain the effect of the semantic outlier features. The $\delta$ is a hyper-parameter of Huber loss (we set $\delta = 1$). Owing to this robust kernel, the label lived in edge can be more exact. We show the comparison results in the ablation study in Sec.~\ref{exp:ablation study}.

The overall loss function is finally formulated as the following minimization problem:
\begin{equation}
\begin{aligned}
    \underset{\mathcal{P}}{\operatorname{min}} \, \, \lambda_1\mathcal{L}_{rgb}(\mathbf{r}) &+ \lambda_2\mathcal{L}_{depth} (\mathbf{r}) + \lambda_3\mathcal{L}_{occ}(\mathbf{r})
    + \lambda_4\mathcal{L}_{fs}(\mathbf{r}) \\ &+ \lambda_5\mathcal{L}_{sg}(\mathbf{r}),
\end{aligned}
\end{equation}
where $\mathcal{P} = \{\theta, \phi, \omega, \alpha, \beta, \gamma, \mathbf{\xi}_i\}$ is the list of parameters being optimized, including fields feature, decoders, and camera pose. The weights of each loss item are set $\lambda_1$ = 10.0, $\lambda_2$ = 1.0, $\lambda_3$ = 10.0, $\lambda_4$ = 1.0, $\lambda_5$ = 2.0.

\subsection{Zero-shot Inference}
To achieve open vocabulary 3D understanding for robotics visual navigation tasks, we combine the semantic field and occupancy field $\mathcal{Q}(\cdot)$ into a new explicit feature map under a fixed resolution.
\begin{equation}
    \mathcal{Q}(\mathbf{x}_i) = \mathcal{V}_\theta(\mathbf{x}_i) \otimes \mathcal{S}_\omega(\mathbf{x}_i),
\end{equation}
where $\otimes$ denotes an operation that selects the semantic feature in which the occupancy probability is 1 at position $\mathbf{x}_i$. Given a set of language tokens $\mathbf{t}_i \in \mathcal{T}$, we transfer these words into prompts and utilize a text encoder $\mathcal{E}$ to obtain text features. The final segmentation results for each 3D point $\mathbf{x}_i$ can be computed point-wise by:
\begin{equation}
    s(\mathbf{x}_i) = \operatorname{argmax}_i\{\cos(h_\gamma\left(\mathcal{Q}(\mathbf{x}_i)\right), \mathcal{E}(\mathbf{t}_i))\}
\end{equation}

\section{Experiment}
\label{sec:exp}
\subsection{Experimental Setup}
\noindent \textbf{Datasets.} 
We train and evaluate our model mainly on three popular public benchmarks: Replica~\cite{Replica}, ScanNet~\cite{dai2017scannet}, and Matterport3D~\cite{Matterport3D_2017}. These three datasets encompass the major situations of indoor scenes: the first two represent regular indoor scenes, and the last represents complex and large-scale indoor scenes, which all provide a series of posed RGB-D images. To more accurately and authentically showcase the segmentation capabilities of these methods on previously unaddressed classes, we also evaluate our method on the Scannet-200 dataset~\cite{rozenberszki2022language}.

\noindent \textbf{Metrics.} 
The evaluation of the reconstructed 3D structure involves assessing six standard metrics as defined in~\cite{murez2020atlas}: Accuracy (Acc), Completeness (Comp), Chamfer-$\mathcal{L}_1$ (C-$\mathcal{L}_1$), Precision (Prec), Recall, and F-score. The F-score is computed using a threshold of 5cm. For scene understanding, we compute the two most common evaluation metrics, mean intersection-over-union (mIou) and mean accuracy (mAcc), which are used to evaluate the segmentation performance and accuracy of models. There is no unified and accurate evaluation metric for open vocabulary scene perception, so we mainly perform qualitative comparisons and discussions to demonstrate the zero-shot capability of our proposed method. 

\noindent \textbf{Implementation Details.} 
We run our OpenOcc on a Tesla A6000 GPU for about 30 to 40 minutes(depending on the scene size) to finish the reconstruction and perception, which takes roughly 30GB of memory in total. For all our experiments,  we sample 6144 rays per patch and sample 132 points along each ray.  We further train our model for 10k iterations with a learning rate of $1 \times 10^{-2}$ and $1 \times 10^{-3}$ for MLP decoders and feature grids. Due to the universality of our approach, unconstrained by specific semantic segmentation methods, we choose pre-trained OpenSeg ~\cite{gu2022openvocabulary}  for 2D feature distillation in our experiment part.

\subsection{Experimental Results}
\label{subsec:exp_rel}

\begin{table}[!t]
\caption{\textbf{Classwise semantic segmentation results on the ScanNet-200~\cite{rozenberszki2022language} validation set.} We show the quantitative comparison between our method and the most recent zero-shot 3D segmentation approach: 
OpenScene~\cite{peng2023openscene}.}
\vspace{-2mm}
    \centering
    \setlength{\tabcolsep}{0.12cm}
            \begin{tabular}{l|cccc|cccc}
            \toprule
             &  \multicolumn{4}{c|}{mIoU} & \multicolumn{4}{c}{mAcc} \\ 
            {} & Clock & Light & Fan & Book & Clock & Light & Fan & Book\\
            \midrule
            OpenScene
            &37.6 & 1.7 & 6.4 &1.0 & 37.7 & 53.8 & 25.9 &1.0\\
            \textbf{Ours}& \textbf{52.9} & \textbf{8.0} & \textbf{29.9} &\textbf{19.6} & \textbf{66.1} & \textbf{75.9} & \textbf{32.8} &\textbf{25.9}\\
            \bottomrule
            \end{tabular}
\label{tab:class}
\vspace{-2mm}
\end{table}
\begin{table}[!t]
    \caption{\textbf{Comparisons on Indoor Benchmarks.}  We compare our method with the zero-shot baselines for 3D semantic segmentation on Indoor Benchmarks.}
    \vspace{-2mm}
    \centering
    \footnotesize
    \setlength{\tabcolsep}{0.08cm}
        \begin{tabular}{l|cc|cc|cc}
            \toprule
            {} & \multicolumn{2}{c|}{Replica~\cite{Replica}} & \multicolumn{2}{c|}{ScanNet-200~\cite{rozenberszki2022language}} & \multicolumn{2}{c}{Matterport~\cite{Matterport3D_2017}}\\
            {} &mIoU & mAcc& mIoU & mAcc& mIoU & mAcc\\\hline
            MSeg Voting~\cite{MSeg_Lambert_2020}  & - & - & - & - & 33.4 & 39.0 \\
            OpenScene~\cite{peng2023openscene}  & 46.9 & \textbf{69.0} &13.3 & 25.7 & 42.5 & 60.4\\
             \textbf{Ours}  & \textbf{50.5} & 65.4 & \textbf{17.5} & \textbf{26.8} & \textbf{45.1} & \textbf{63.3}\\
            \bottomrule
        \end{tabular}
\label{tab:benchmark}
\vspace{-3mm}
\end{table}

\paragraph{Results on 3D semantic segmentation} 
We evaluate our OpenOcc on all classes of the Replica, Scannet-200 validation set and Matterport3D test set. In Tab.~\ref{tab:class} and Tab.~\ref{tab:benchmark}, we compare our methods with the most closely related work on zero-shot methods.
Due to the original LeRF approach not utilizing depth supervision, it inherently suffers from shape-radiance ambiguity. Consequently, we have not compared our method with it in this part.
Our model integrates 3D scene reconstruction with open vocabulary segmentation, exhibiting superior performance relative to OpenScene, particularly in the segmentation of small objects or long-tail objects, as evidenced in Tab. \ref{tab:class}. 
Since reconstructing a large number of scenes will consume too much resource and time, we establish a new validation subset by randomly extracting 10\% of data(about 30 scenes) from the Scannet validation set, and a similar operation is performed on the Matterport3D test set. For fairness, we also evaluate OpenScene under the same subset split. 

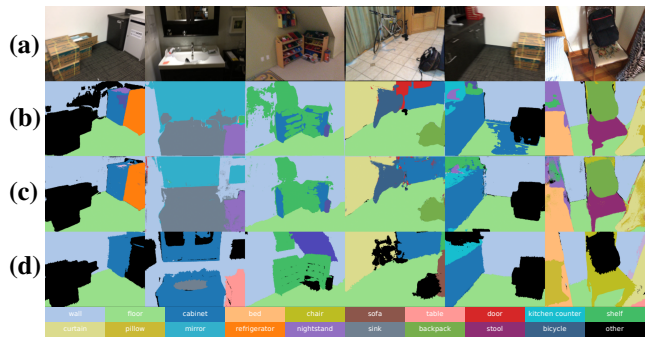
\begin{figure}[t]
\flushright
\includegraphics[width=0.96\linewidth]{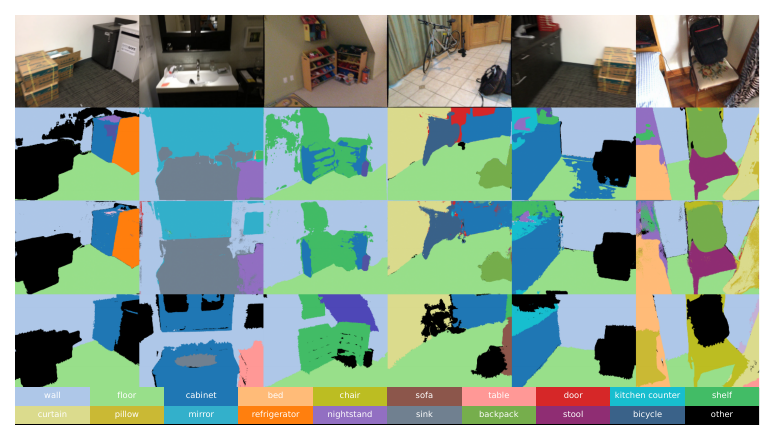}
\put(-245,112){{\bf{(a)}} }
\put(-245,84){{\bf{(b)}} }
\put(-245,56){{\bf{(c)}} }
\put(-245,28){{\bf{(d)}} }
\vspace{-2mm}
\caption{\textbf{2D segmentation results on ScanNet.}
We visualized some 2D segmentation examples from the ScanNet validation set. (a) depicts the Input Image, while (b) showcases the OpenSeg result, (c) illustrates our method, and (d) represents GT Segmentation. Black pixels in the ground truth segmentation correspond to classes not included in the ScanNet-20 evaluation classes. }
\label{fig:scannet2D}
\vspace{-3mm}
\end{figure}

As shown in Tab.~\ref{tab:benchmark}, our model outperforms the zero-shot baselines (OpenScene) in terms of mIou and mAcc on Matterport3D and ScanNet-200 and achieves a higher mIou score in Replica. Owing to the motion blur and invalid depth value of collected images on ScanNet, it is a challenge for most reconstruction methods and will significantly reduce the reconstruction quality. In contrast, OpenScene extracts the point cloud from the GT mesh, which undergoes post-processing operations. This leads to variances in the quality of the reconstructed geometry. However, we demonstrate that our method still achieves superior results in classes previously unaddressed, as evidenced by our findings on the ScanNet-200 dataset presented in Tab. \ref{tab:class}. Owing to the SCP method relieving the issue of language field representation degeneracy, our approach exhibits enhanced segmentation capability for small objects, including vases, pillows, and stools, as illustrated in Fig. \ref{fig:occ_seg}. This finding aligns with the conclusions presented in Tab. \ref{tab:class}.

\paragraph{Results on 2D semantic segmentation} 
Compared to other 3D segmentation methods, our method can also obtain any view of 2D segmented results owing to the view synthesis capabilities of NeRFs. 
Visual comparisons of 2D semantic segmentations on ScanNet are shown in Fig. ~\ref{fig:scannet2D}. Despite errors or inconsistencies in the 2D segmentation from OpenSeg, our method can still correctly segment and complete the targets in the scene through volume rendering. For example, the OpenOcc successfully segmented the wall in the first column of Fig.~\ref{fig:scannet2D} and completely identified the floor in the fifth column of Fig.~\ref{fig:scannet2D}. This demonstrates that our method 
can better fuse the language embeddings and obtain more consistent, complete, and accurate 2D segmentation results. 

\begin{table}[!t]
    \caption{ \textbf{Quantitative evaluation of reconstruction map.} 
    We evaluate the 3D reconstruction performance of our proposed method against previous works on ScanNet dataset with six represent metrics. ``$*$" denotes the results generated from the source code with the default parameters.}
    \vspace{-2mm}
    \centering
    \footnotesize
    \setlength{\tabcolsep}{0.1cm}
        \begin{tabular}{l|cccccc}
            \toprule
            Method & Acc$\downarrow$ & Comp$\downarrow$ & C-$\mathcal{L}_1\downarrow$ &Prec$\uparrow$ &Recall$\uparrow$ & F-score$\uparrow$ \\ \hline
            COLMAP~\cite{schoenberger2016mvs} &0.047  &0.235  &0.141   &71.1  &44.1   &53.7  \\
            UNISURF~\cite{oechsle2021unisurf} &0.554  &0.164  &0.359   &21.2  &36.2   &26.7  \\
            VolSDF~\cite{yariv2021volume} &0.414  &0.120  &0.267   &32.1  &39.4   &34.6  \\
            NeuS~\cite{wang2021neus} &0.179  &0.208  &0.194   &31.3  &27.5   &29.1  \\
            Manhattan-SDF~\cite{guo2022manhattan} &0.072  &0.068  &0.070   &62.1  &56.8   &60.2  \\
            NeuRIS~\cite{wang2022neuris} &0.050  &0.049  &0.050   &71.7  &66.9   &69.2  \\
            MonoSDF~\cite{MonoSDF} &\textbf{0.035} &0.048  &0.042   &79.9  &68.1   &73.3  \\
            OccSDF~\cite{Occ-SDF} &0.039  &0.041  &0.040   &80.0  &76.0   &77.9  \\
            Go-Surf*~\cite{wang2022go} &0.102  &\textbf{0.024}  &0.063   &79.0  &\textbf{91.2}   &84.2  \\
            \textbf{Ours} &0.044  &0.029 &\textbf{0.037}  &\textbf{81.2}  &90.0   &\textbf{85.3} \\
            \bottomrule
        \end{tabular}
\label{tab:recon}
\end{table}
\begin{figure}[t]
	\centering
	\begin{minipage}{0.24\linewidth}
		  \centering
  \begin{tikzpicture}
  \node[anchor=south west, inner sep=0] (image) at (0,0) {\includegraphics[width=1.0\textwidth]{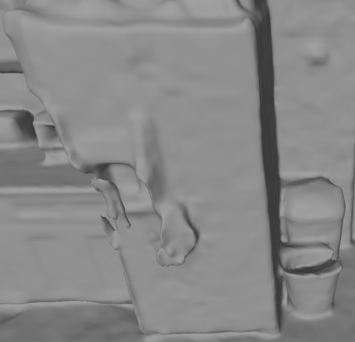}};
  \begin{scope}[x={(image.south east)},y={(image.north west)}]
  \draw[red, thick, dashed] (0.6,0.7) rectangle (0.2,0.2);
  \end{scope}
  \end{tikzpicture}
	\end{minipage}
 	\begin{minipage}{0.24\linewidth}
  \centering
  \begin{tikzpicture}
  \node[anchor=south west, inner sep=0] (image) at (0,0) {\includegraphics[width=1.0\textwidth]{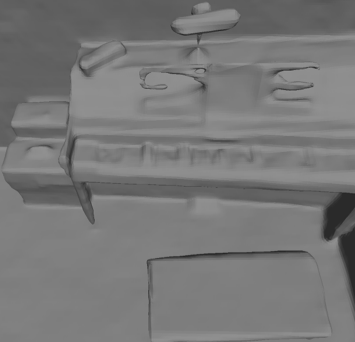}};
  \begin{scope}[x={(image.south east)},y={(image.north west)}]
  \draw[red, thick, dashed] (0.9,0.9) rectangle (0.18,0.45);
  \end{scope}
  \end{tikzpicture}
	\end{minipage}
	\begin{minipage}{0.24\linewidth}
   \centering
  \begin{tikzpicture}
  \node[anchor=south west, inner sep=0] (image) at (0,0) {\includegraphics[width=1.0\textwidth]{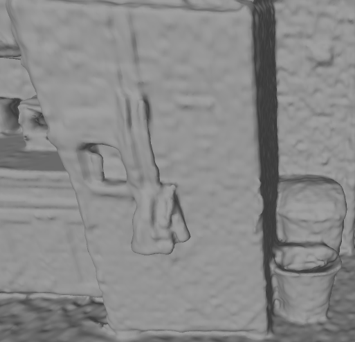}};
  \begin{scope}[x={(image.south east)},y={(image.north west)}]
  \draw[red, thick, dashed] (0.6,0.7) rectangle (0.2,0.2);
  \end{scope}
  \end{tikzpicture}
	\end{minipage} 
	\begin{minipage}{0.24\linewidth}
  \centering
    \begin{tikzpicture}
  \node[anchor=south west, inner sep=0] (image) at (0,0) {\includegraphics[width=1.0\textwidth]{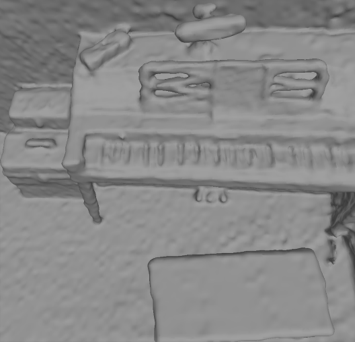}};
  \begin{scope}[x={(image.south east)},y={(image.north west)}]
  \draw[red, thick, dashed] (0.9,0.9) rectangle (0.18,0.45);
  \end{scope}
  \end{tikzpicture}
	\end{minipage}
 
\vspace{3pt}

 \begin{minipage}{0.45\linewidth}
\centering
    \textbf{\small{Go-Surf}}
\end{minipage}
\begin{minipage}{0.45\linewidth}
\centering
    \textbf{\small{Ours}}
\end{minipage}

\vspace{-2mm}
\caption{\textbf{Qualitative results of detailed reconstructed architectures.}  Compared with our baseline Go-Surf~\cite{wang2022go}, our approach demonstrates superior accuracy in reconstructing shapes and contours of objects on Scannet scene 0 (left) and scene 50 (right).
}
\label{fg:reconstruction_cmp}
\vspace{-5mm}

\end{figure}
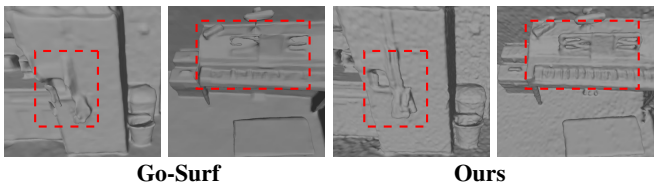

\paragraph{Reconstruction Quality} To better demonstrate the performance of our proposed occupancy representation in scene reconstruction, we evaluate the reconstructed 3d structure from both qualitative and quantitative perspectives on ScanNet. We use the test split from~\cite{guo2022manhattan} and uniformly sample roughly 120 frames to train our network. In our framework, as we generate an occupancy grid, we extract meshes based on the occupancy probability using marching cubes with a resolution of 1cm and a threshold of $0.5$. Regions that are not observed in any camera views are culled before
evaluation. In Tab.~\ref{tab:recon}, we conducted a comparison of our proposed approach against existing implicit reconstruction methods, including Manhattan-SDF~\cite{guo2022manhattan}, MonoSDF~\cite{MonoSDF}, Go-Surf~\cite{wang2022go} and OccSDF~\cite{Occ-SDF}. Our approach achieves comparable performance in 3D scene reconstruction. Compared with our baseline method Go-Surf~\cite{wang2022go}, our approach demonstrates high-quality reconstructions of shapes and contours for objects, as illustrated in Fig.~\ref{fg:reconstruction_cmp}.

\subsection{Application of robotic visual navigation }
\begin{figure}[t]
\centering
\includegraphics[width=0.98\columnwidth]{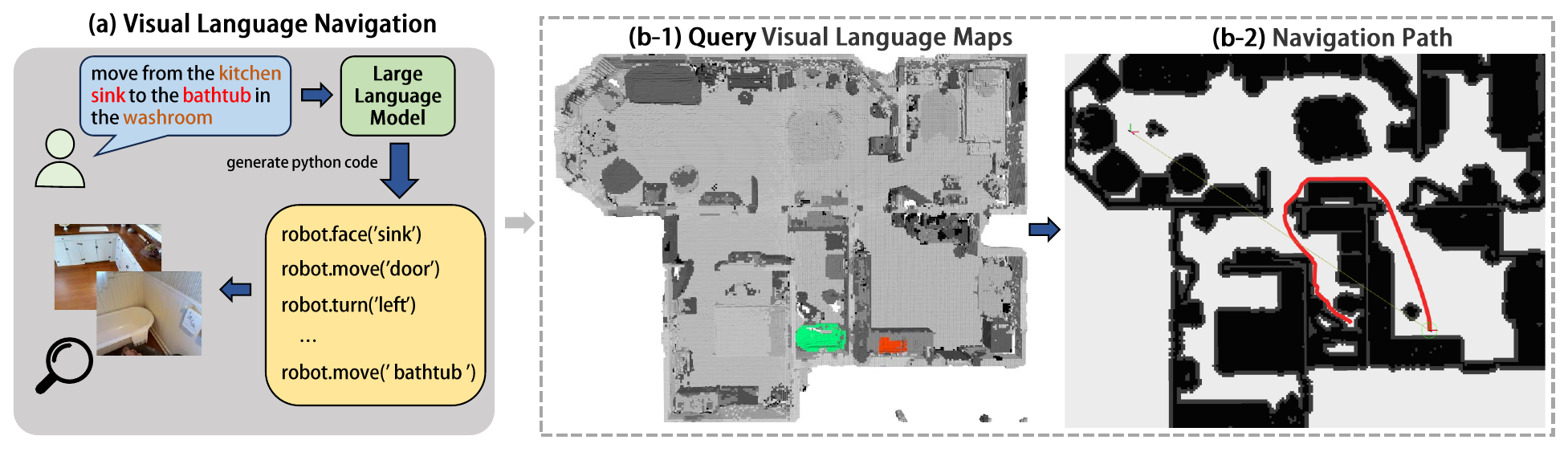}
\vspace{-2mm}
\caption{\textbf{The simulation experiments in robotic visual navigation.}}
\label{fg:sim}
\vspace{-4mm}
\end{figure}

To further illustrate the potential applicability and efficiency of OpenOcc within the area of mobile robotics, we have conducted a simulation experiment for visual navigation tasks. This experiment is designed to highlight the advantages of the proposed method for practical robotic navigation, as evidenced in Fig.~\ref{fg:sim}. Upon receiving navigation instructions, leveraging ChatGPT finishes task decomposition, which aids in acquiring detailed information about the origin (kitchen sink) and destination objects (washroom bathtub). By querying our visual language map, we can obtain precise location information, as illustrated in part b-1 of Fig.~\ref{fg:sim}. Subsequently, a navigation route is devised utilizing the path planning module within ROS2, as depicted in part b-2 of Fig.~\ref{fg:sim}. The open-vocabulary occupancy map we have reconstructed facilitates mobile robots in more effectively completing visual navigation tasks.

\subsection{Memory and Time}
We exhibit the merits of our occupancy-based reconstruction and representation method in terms of low storage requirements and high efficiency, shown in Fig.~\ref{fg:intro query show}. The occupancy-based representation allows for the efficient exclusion of invalid sampled spaces. Compared with the SDF-based NeRF method, our occupancy-based NeRF variant reduces training time by a significant 74.7\%, from 33min down to just 8min. Moreover, implicit fields typically require 4.4GB of memory, whereas our explicit language feature occupancy grid takes only 1.1GB of memory, a reduction of 75\%. 

\begin{table}[!t]
    \caption{
        \textbf{Ablation Study.} Comparison of semantic segmentation performance of different proposed components by our method.
        }
    \vspace{-2mm}
    \centering
    \footnotesize
    \setlength{\tabcolsep}{0.55cm}
        \begin{tabular}{l|c|cc}
            \toprule
             {} & mIoU & mAcc \\\hline
            w/o Huber& 44.6  & 62.6  \\
            w/o SCP  &  43.7 &  62.8 \\
            w/o BCE & 44.0 &   59.6 \\
            OpenOcc  &  \textbf{45.1}  &   \textbf{63.3}\\
            \bottomrule
        \end{tabular}
\label{tab:ablation_feature}
\end{table}

\subsection{Ablation Study}
\label{exp:ablation study}
To verify the effectiveness of our network design, including the Huber loss, Binary Cross Entropy (BCE) loss, and the Semantic-aware Confidence Propagation (SCP) module, we test the performance of our method under different settings.
The results presented in Tab.~\ref{tab:ablation_feature} demonstrate that all the proposed strategies are effective on matterport3D segmentation. This suggests that adopting the BCE loss can promote better occupancy-based reconstruction. At the same time, leveraging  SCP and the robust kernel can effectively eliminate training instability caused by inconsistent 2D open segmentation.

\section{Conclusion}
\label{sec:conclusion}
We present OpenOcc, a novel framework that integrates 3D scene understanding and reconstruction using Neural Radiance Fields, which combines the advantage of the computationally efficient occupancy representation and the open vocabulary scene understanding capability. Experimental results demonstrate our method can reconstruct high-quality geometric structures and exhibit competitive performance on three public benchmarks compared with current state-of-the-art scene understanding methods. The supplementary external simulation experiment further underscores the prospective applicability of our proposed method for robotic visual navigation.

\bibliographystyle{IEEEtran}
\bibliography{root}

\end{document}